\documentclass[letterpaper, 10 pt, conference]{ieeeconf} 

\IEEEoverridecommandlockouts               

\usepackage{graphics} 
\usepackage{epsfig} 
\usepackage{algorithm}
\usepackage{algpseudocode}

\title{\LARGE \bf
A Hybrid Control Design for Autonomous Vehicles at Uncontrolled Intersections}

\author{Nitin R. Kapania$^{1}$, Vijay Govindarajan$^{2}$, Francesco Borrelli$^{2}$, J. Christian Gerdes$^{1}$
\thanks{*This work was supported by the Hyundai Center of Excellence}
\thanks{$^{1}$Dept. of Mechanical Engineering, Stanford University. Nitin Kapania is also a visiting scholar at UC Berkeley.}%
\thanks{$^{2}$Dept. of Mechanical Engineering, UC Berkeley.}%
}

\begin{document}

\maketitle
\thispagestyle{empty}
\pagestyle{empty}

\begin{abstract}

As autonomous vehicles (AVs) inch closer to reality, a central requirement for acceptance will be earning the trust of humans in everyday driving situations. In particular, the interaction between AVs and pedestrians is of
high importance, as every human is a pedestrian at some point of the day. This paper considers the interaction of a pedestrian and an autonomous vehicle at a mid-block, unsignalized intersection where there is ambiguity
over when the pedestrian should cross and when and how the vehicle should yield. By modeling pedestrian behavior through the concept of gap acceptance, the authors show that a hybrid controller with just four distinct modes allows an autonomous vehicle to successfully interact with a pedestrian across a continuous spectrum of possible crosswalk entry behaviors. The controller is validated through extensive simulation and compared to an alternate POMDP solution and experimental results are provided on a research vehicle for a virtual pedestrian. 

\end{abstract}

\section{Introduction}

\subsection{Motivation}

While autonomous vehicles have the potential to save thousands of lives every year and create significant societal benefits \cite{Fagnant2015}, widespread adoption is unlikely until AVs gain the broad trust of society. Given that every human is a pedestrian at some point during the day, one of the central ways that autonomous vehicles will be evaluated is through their interactions with pedestrians. Interactions with pedestrians can be complex, even for experienced human drivers. From 2015-2016, pedestrian fatalaties increased by 9\% to 5987, representing the highest number since 1990, and also representing 16\% of all automotive fatalities \cite{HighwayTrafficSafetyAdministration2016}. As autonomous vehicles inch closer to widespread adoption, they must have a clear control strategy for pedestrian interaction that can handle a wide variety of pedestrian behaviors while maintaining a reasonable flow of traffic.  

\subsection{Prior Art}

The need to further understand pedestrian behavior for autonomous driving has created a signficant body of research focused on modeling pedestrian behavior given various sensor inputs. Keller et al. \cite{Keller2014} presented a study on pedestrian path prediction and action classification (e.g. crossing vs. waiting) using Gaussian process dynamical models and trajectory matching from optical data. Several other techniques for pedestrian trajectory prediction have been proposed, including LQR \cite{Batkovic}, set-based reachability analysis \cite{Koschi2018}, and Markov processes \cite{Karasev2016}. Another branch of literature focuses on modeling pedestrian crossing behavior in the form of video analysis. Schroeder and Rouphail \cite{Schroeder2011} explored factors associated with driver yielding behavior at unsignalized pedestrian crossings. Using logistic regression, the authors found that drivers are more likely to yield to assertive pedestrians who walk briskly in their approach to a crosswalk. Kadali and Perumal \cite{RaghuramKadali2012} studied the ``gap acceptance" behavior of pedestrians at mid-block crosswalks, and found that the gap accepted for crossing was explained by factors such as crossing direction, vehicle speed, and pedestrian age. Yannis et al. \cite{Yannis2013} and Sun et al.\cite{Sun2002} also found that gap acceptance was influenced by the size of the oncoming vehicle and the presence of other pedestrians. 

A small but rapidly growing body of literature specifically studies the interaction between pedestrians and autonomous vehicles at crosswalks. An excellent review of these studies was conducted by Rasouli et al. \cite{Rasouli}. As an example, Rothenbucher et al. \cite{Rothenbucher2016} studied the interaction between pedestrians and driverless vehicles by constructing a car seat costume to disguise a driver. To improve the issue of trust, several researchers have also developed external communication interfaces to more clearly broadcast the intent of the autonomous vehicle (\cite{Matthews}, \cite{Lagstrom2015}). 

One issue with automated driving for crosswalk scenarios is that an overly conservative crossing algorithm will often be taken advantage of or cause confusion among pedestrians\cite{Camara2018}. Camara et al. \cite{Camara2018} attempted to model the natural negotiation for priority between a pedestrian and an AV at an intersection using the framework of game theory. Chen et al. \cite{Chen} notes the need for a tradeoff between passive and aggressive driving behavior, and developed a stochastic model of pedestrian behavior to evaluate proposed AV control policies. Control polices for pedestrian interaction are also developed in \cite{Bandyopadhyay}, who proposes a Mixed-observable Markov Decision Process (MOMDP) to incorporate the intention uncertainty of a pedestrian. A Partially-observable Markov Decision Process (POMDP) formulation is also proposed by Thornton \cite{Thornton2018}, who proposes a value-sensitive design framework to assist with the development of engineering specifications for the control algorithm. 

\subsection{Statement of Contributions}
Significant research has been conducted to understand the likelihood of pedestrian crossing given certain traffic conditions and pedestrian demographics \cite{Schroeder2011} -\cite{Lee2005}, and a small but growing body of literature develops control strategies for pedestrian avoidance \cite{Bandyopadhyay}-\cite{Thornton2018}. However, the literature lacks a contribution that combines the two and explicitly tests whether a proposed control strategy is robust to the variety of pedestrian behaviors that have been observed from experimental studies on real roads. What is also lacking is analysis of how this controller should behave across multiple traffic scenarios - for example, the navigation problem is different for the pedestrian crossing on the opposing side of traffic. To the authors knowledge, there are also no studies that explicitly compare different approaches in order to provide a comparision of alternative solution methods. 

This paper aims to address the above issues by developing a hybrid control architecture that accounts for several distinct pedestrian modes of behavior at an unsignalized crosswalk. Simulations show that the proposed hybrid controller is able to handle a continuous spectrum of pedestrian \textit{gap acceptance} behavior, tolerating a range of highly conservative to highly aggressive pedestrians. Additionally, this paper provides a simulated comparison between the hybrid controller and the solution method proposed by Thornton \cite{Thornton2018}, citing advantages and disadvantages of each method. Finally, experimental results are shown on a real vehicle to demonstrate the feasibility of the proposed controller.

\section{Problem Overview}

\subsection{Unsignalized Intersections}

\begin{figure}
\centering
\includegraphics[width=2.5in]{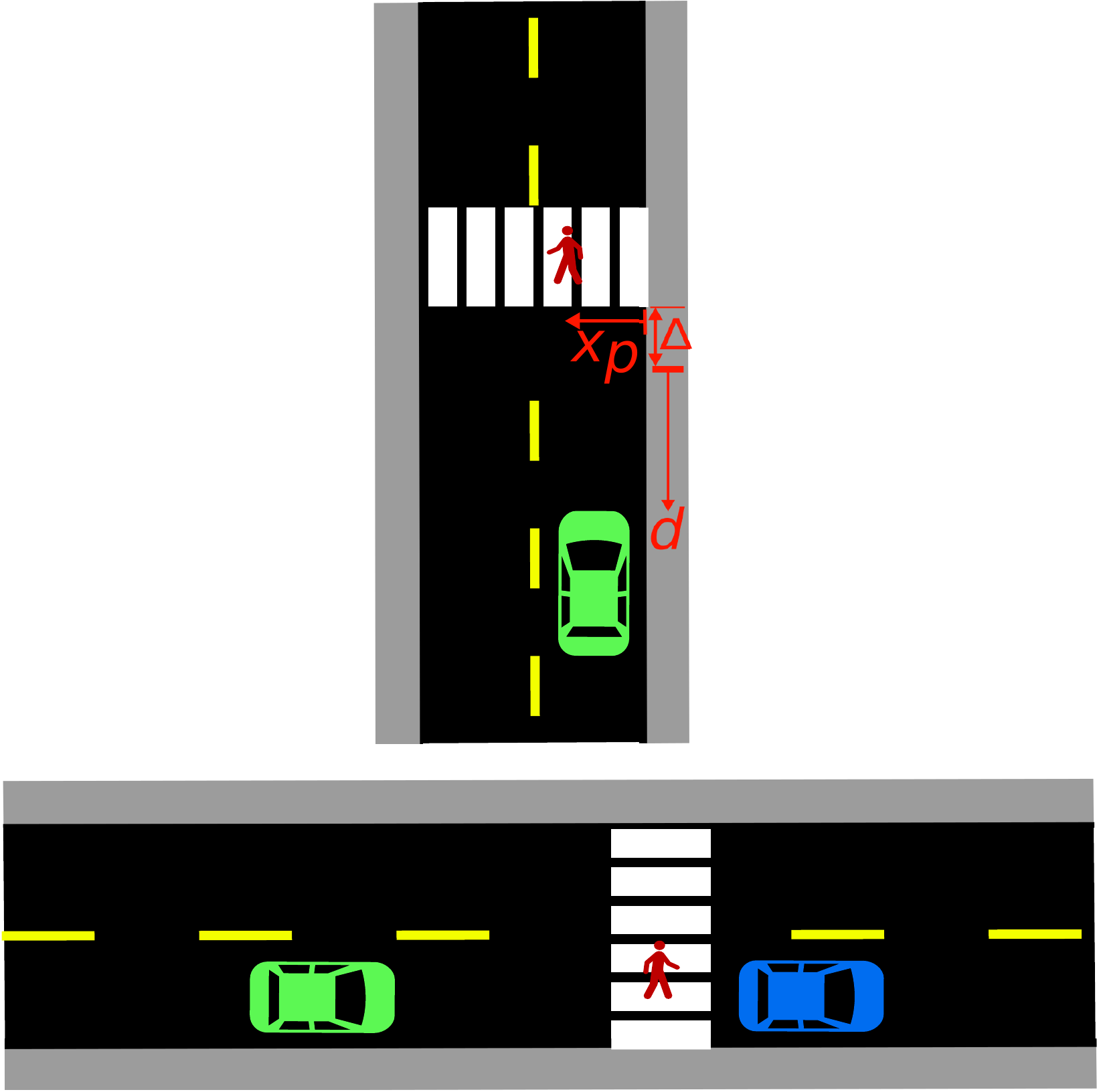}
\caption{Top: Description of relevant problem variables. Bottom: Schematic of pedestrian approaching a stream of traffic at a mid-block intersection. Ego vehicle shown in green. We develop a hybrid control framework for the green vehicle, which considers the distance ($d$) to a stopping point $\Delta$ ahead of the crosswalk and the pedestrian's position in the crosswalk ($x_p$).}
\label{fig:schematic}
\end{figure}

In general, pedestrian crosswalks may be controlled or uncontrolled. In the case of the former, control devices such as stop lights and walk signals guide the interaction of vehicles and pedestrians explicitly. Fig.~\ref{fig:schematic} shows the latter example, in which there is no such control device, and a pedestrian must select a gap in traffic flow and cross.  

In general, right-of-way for uncontrolled intersections is complex. For example, nine states and the District of Columbia require motorists to \textit{stop} when approaching a pedestrian in an uncontrolled crosswalk, while another nineteen states only require a motorist to \textit{yield} when a pedestrian is upon any portion of the roadway. A full guideline for right-of-way for all fifty states is provided in \cite{Legislatures2018}.

\subsection{Problem Formalization}
\label{sec:probform}

A diagram of the relevant state variables is shown in Fig.~\ref{fig:schematic}. We assume that knowledge of distance $d(t)$ to a fixed stopping point $\Delta$ 3-5 meters ahead of the crosswalk can be estimated by a perception system, and that the vehicle velocity $v(t)$ is known as well. We also assume the perception system is able to determine the position of the pedestrian $x_p$ within the crosswalk, and is able to determine a velocity estimate $\dot{x}_p$ for the pedestrian. Note that $x_p$ can be negative for the case where the pedestrian is on the sidewalk. 


The relevant control problem is therefore to determine an appropriate longitudinal acceleration command for the vehicle, $u(t) = a(t)$, as a closed-loop function of the vehicle and pedestrian states:

\begin{equation}
a = k(d, v, x_p, \dot{x_p})
\end{equation}


\subsection{System Requirements}

A difficulty with this control problem is that there are multiple stakeholders and therefore multiple competing objectives that must be considered. Relevant stakeholders for this problem are the autonomous vehicle occupants, the pedestrian(s), and the other vehicles in the traffic stream. 

To formalize the system requirements for this multi-stake holder problem, we propose the following \textit{design requirements}, extended from the engineering specifications proposed by Thornton \cite{Thornton2018}. 

\begin{center}
\begin{tabular}{ l| l }
\label{tb:specs}
 \textbf{Engineering} & \textbf{Design} \\
 \textbf{Specification\cite{Thornton2018}} & \textbf{Constraint}\\\hline 
 1. Safety and Legality & a. Avoid collisions for all\\
           & possible accepted gaps            \\
           &b. Ability to follow stop/yield  \\
           &laws for all 50 U.S. states\\
           
&\\
 2. Efficiency & a. Average speed: Minimize \\ 
        & deviation from speed of traffic \\  
        & b. Wait for pedestrian crossing \\
        & event to materialize before \\
        & yielding. \\ 

        & \\
 3. Smoothness & a. Low acceleration: Nominal \\ 
 			  & brake and acceleration limits \\
 			  & of 2 $\mathrm{m/s^2}$ \\

\end{tabular}
\end{center}

\subsection{Pedestrian Gap Acceptance}

One of the major factors that determines pedestrian crossing behavior at uncontrolled intersections is \textit{gap acceptance}, defined in this paper as:

\begin{equation}
\mathrm{gap} = \frac{\mathrm{distance\hspace{1mm}to\hspace{1mm}crosswalk}}{\mathrm{vehicle\hspace{1mm}speed}} = \frac{d}{v}
\end{equation} 

The accepted gap is a measure of how much time there is before the vehicle would enter the crosswalk if it kept its current speed constant. When faced with a stream of traffic at an unsignalized crosswalk, a pedestrian inherently decides how much of a time gap to accept before crossing. The average gap acceptance is reported in the literature to be between 3-7 seconds, meaning pedestrians usually do not cross if the vehicle would enter the crosswalk in under three seconds \cite{DiPietroCharlesMandKing1970}, and are very likely to cross when they have more than seven seconds \cite{Schmidt2009}. 

\section{Proposed Control Architecture}

\begin{figure}
\centering
\includegraphics[width=3.5in]{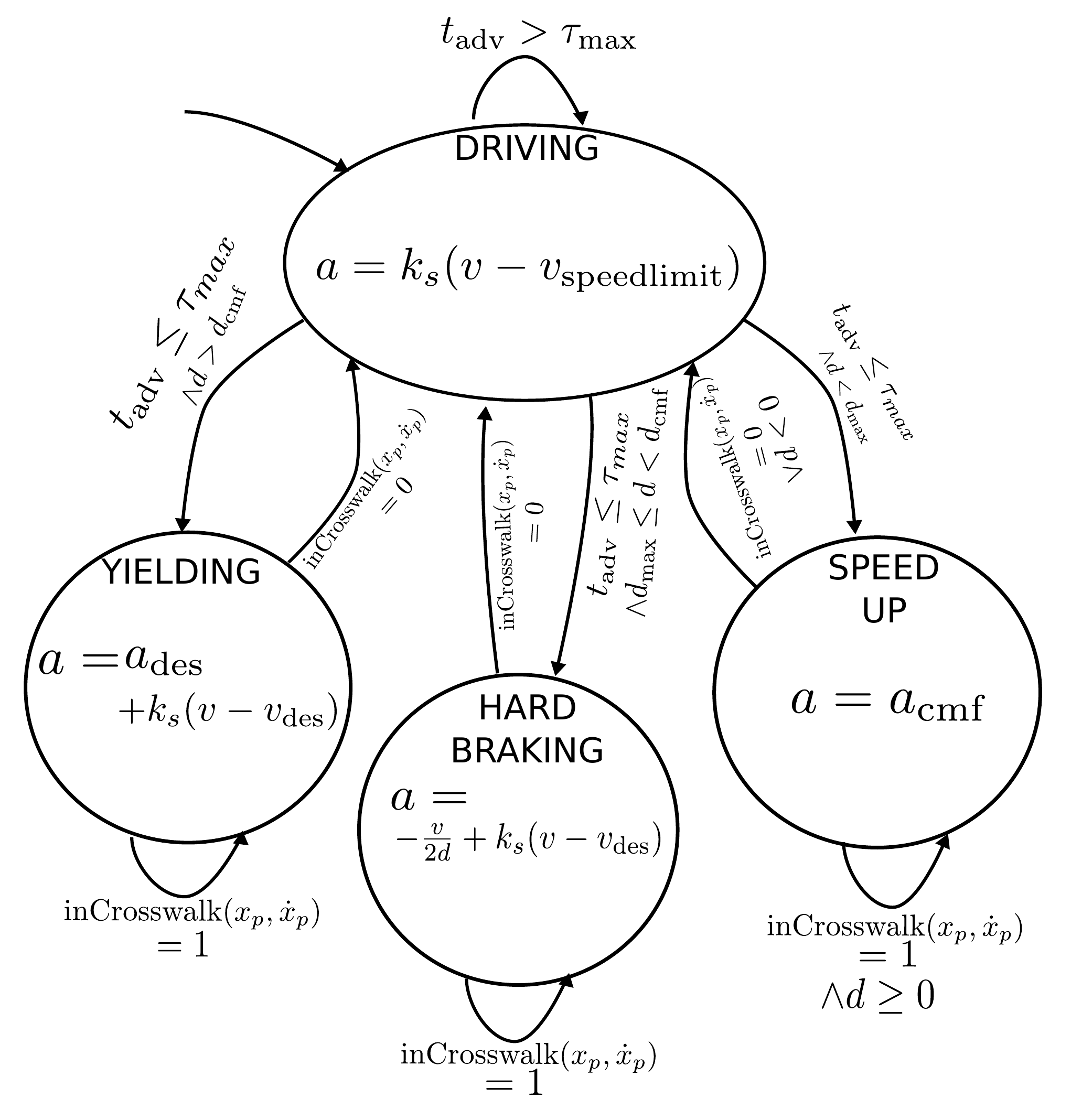}
\caption{Schematic of hybrid control structure.}
\label{fig:hybridController}
\end{figure}

The proposed hybrid control architecture is shown in Fig.~\ref{fig:hybridController}, and the algorithm for operation is shown in Algorithm \ref{alg:hybridController}. At its core, the control algorithm uses a feedback-feedforward methodology to compute desired acceleration commands, but computes these commands differently depending on which of four discrete modes the controller is in. The next section will describe the four modes. 

\begin{table}[h]
\begin{center}
\caption{Parameters for Algorithm 1}\label{tb:params}
\begin{tabular}{lc}
Symbol & Definition \\\hline\hline
$d_\mathrm{max}$ & $\frac{v^2}{2a_\mathrm{max}}$ \\
$d_\mathrm{cmf}$ & $\frac{v^2}{2a_\mathrm{cmf}}$ \\\hline
\end{tabular}
\end{center}
\end{table}

\begin{algorithm} 
\caption{Calculate $a = k(d, v, x_p, \dot{x}_p)$} 
\label{alg:hybridController} 
\begin{algorithmic}
\State $state \leftarrow DRIVING$
\While {$\mathbf{true}$}
\If {$state = DRIVING$}
	\State $a \gets k_s(v-v_\mathrm{speed limit})$ 
	\If{$d > 0 \hspace{1mm} \mathbf{and} \hspace{1mm} \mathrm{inCrosswalk}(x_p)$}
		\If{$\mathrm{timeAdvantage}(x_p, \dot{x_p}, d, v) > \tau_\mathrm{max}$} 
		\State $state \leftarrow DRIVING$
		
		\ElsIf{$d > d_\mathrm{cmf}$}
		\State $state \leftarrow YIELDING$


 \ElsIf{$d_\mathrm{max} < d < d_\mathrm{cmf} $}

	\State $state \leftarrow HARD\hspace{1mm}BRAKING$

	\Else
	\State $state \leftarrow SPEED\hspace{1mm}UP$

	\EndIf
	\EndIf
	\EndIf
\vspace{3mm}
\If {$state = YIELDING$}
\If {$d > d_\mathrm{cmf} + t_{delay}v$}
	\State $a_{des} = 0$
	\State $v_{des} = v_\mathrm{speed limit}$
\Else 
	\State $a_{des} = -a_\mathrm{cmf}$
	\State $v_{des} = \mathrm{getDesiredSpeed}(d, v)$
\EndIf 
	\State $a = a_{des} + k_s(v - v_{des})$

\If {$\textbf{not} \hspace{1mm} \mathrm{inCrosswalk}(x_p)$} \State $state \leftarrow DRIVING$

\EndIf
\EndIf

\vspace{3mm}

\If {$state = HARD\hspace{1 mm}BRAKING$}
	\State $v_{des} = \mathrm{getDesiredSpeed}(d, v)$
	\State $a = -\frac{v^2}{2d} + k_s(v - v_{des})$ 	
	\If {$\textbf{not} \hspace{1 mm} \mathrm{inCrosswalk}(x_p)$} \State $state \leftarrow DRIVING$

	\EndIf
	\EndIf

\If {$state = SPEED\hspace{1mm}UP$}
	\State $a = a_\mathrm{cmf}$ 	
	\If {$\textbf{not} \hspace{1mm} \mathrm{inCrosswalk}(x_p) \vee d < 0$} \State $state \leftarrow DRIVING$

	\EndIf
	\EndIf

\EndWhile

\end{algorithmic}
\end{algorithm}

\subsection{Nominal Driving}
In the nominal state, the vehicle attempts to drive through the crosswalk at the speed limit $v_\mathrm{speedlimit}$. A proportional speed control is applied to keep the vehicle at the speed limit, achieving design constraint 2.a:

\begin{equation}
	a = k_s(v - v_\mathrm{speedlimit})
\end{equation}

The controller remains in the driving mode unless the pedestrian begins to enter the crosswalk before the vehicle has passed. We mathematically define ``beginning to enter" as the instant the pedestrian begins moving towards the crosswalk, even if the pedestrian is currently on the sidewalk:

\begin{equation}
\label{eq:crossing}
\mathrm{inCrosswalk}(x_p, 
\dot{x}_p) = \left\{
 \begin{array}{ll}
  1&: \dot{x}_p \neq 0 \vee \hspace{1 mm} 0 \leq x_p \leq x_F \\
  0&: \mathrm{otherwise} 
 \end{array}
\right.
\end{equation}

 Where $x_F$ is the end of the crosswalk area of interest. According to design constraint 1.b, the definition of $x_F$ will vary according to which US state the vehicle is in. For example, in California, $x_F$ would be at the end of the crosswalk, while in Louisiana, $x_F$ would just be the same half of the crosswalk as the vehicle \cite{Legislatures2018}. Once the pedestrian enters the crosswalk, the algorithm decides which mode to enter next. 

\subsubsection{Time Advantage}

Frequently the vehicle will pass through the crosswalk well before the pedestrian will reach the vehicle's lane. Assuming the vehicle is operating in a US state where stopping is not explicitly required, design constraint 2.a suggests it is desirable for the vehicle to continue through the intersection.

This can be formalized by calculating the \textit{time advantage} $t_\mathrm{adv}$ \cite{Chen} as follows:

\begin{equation}
t_\mathrm{adv} = \frac{x_v - x_p}{\dot{x_p}} - \frac{d}{v}
\end{equation}

Where $x_v$ is the $x$ position of the vehicle in the crosswalk. Algorithm 1 explicitly allows the vehicle to continue driving through the intersection as long as $t_\mathrm{adv}$ exceeds a specified threshold $\tau_\mathrm{max}$. 

\subsubsection{Yielding}

If the time advantage is not sufficient for the vehicle to pass through, the autonomous vehicle must slow down. The preferred option to meet design constraint 3.a is to brake at a low, comfortable deceleration $-a_\mathrm{cmf}$ for the passenger.

Comfortable braking at a deceleration $-a_\mathrm{cmf}$ is possible if sufficient braking distance exists. The braking distance is derived kinematically as 

\begin{equation}
d_{brake} = \frac{v^2}{2a_\mathrm{cmf}}
\end{equation}

If this holds true when the pedestrian starts to enter the crosswalk, the car enters a \textit{Yielding} mode. 

\subsubsection{Hard Braking}

In the unlikely event the pedestrian crosses when the time gap is low, the vehicle will need to prioritize design constraint 1.a and come to a stop at a deceleration higher than $-a_\mathrm{cmf}$. This occurs when the following conditions holds:

\begin{equation}
\frac{v^2}{2a_\mathrm{max}} < d < \frac{v^2}{2a_\mathrm{cmf}}
\end{equation}

Where $a_\mathrm{max}$ is the largest deceleration magnitude allowed by the tire-road friction. 

\subsubsection{Speeding Up}

As a final condition, consider the case where the pedestrian enters the crosswalk just as the vehicle is crossing $d = 0$. In this case, the vehicle has insufficient space to stop before the crosswalk and will risk being rear-ended or stopping in the crosswalk. It makes more sense in this condition for the vehicle to speed up and exit the crosswalk quickly. As a result, if $d < \frac{v}{2a_\mathrm{max}}$, the vehicle speeds up. 

\subsection{Yielding}

In the yielding state, the vehicle follows a different version of the feedback-feedforward dynamics. The vehicle first follows the speed limit until it reaches the critical value of $\frac{v^2}{2a_\mathrm{cmf}}$. In practice, an additional term $t_\mathrm{delay}v$ is added to compensate for the delay in brake communication. 

At the critical distance, the vehicle decelerates at the uniform yielding deceleration of $-a_\mathrm{cmf}$, with an additional feedback term $k_s(v - v_\mathrm{des}$), where $v_\mathrm{des}$ is the desired velocity. The feedback term helps bring the vehicle speed to 0 at the desired stopping point. 

The desired velocity profile $v_\mathrm{des}(d)$ for a vehicle decelerating at constant acceleration is given by: 

\begin{equation}
v_\mathrm{des}(d) = \sqrt{2a_\mathrm{cmf}(d-d_o) + v_o^2}
\end{equation}

Where $d_o$ and $v_o$ are the values of $d$ and $v$ when the controller first enters the yielding state. Inspection shows that the desired velocity is $v_o$ when $d = d_o$ and is 0 when $d = 0$. 

The controller exits the yielding mode and returns to the driving mode once the pedestrian clears the crosswalk. 

\subsection{Hard Braking}

In the hard braking state, the vehicle follows a similar feedback-feedforward algorithm, but the desired feedforward deceleration in this case is given by $a = -\frac{v^2}{2d}$. Integrating to find the desired speed profile $v_\mathrm{des}(d)$ yields the following velocity profile: 

\begin{equation}
v_\mathrm{des}(d) = \frac{v_o}{\sqrt{d_o}}\sqrt{d}
\end{equation}   

Where $d_o$ and $v_o$ are the values of $d$ and $v$ when the controller first enters the braking state. Again, the controller exits the braking mode and returns to the driving mode once the pedestrian clears the crosswalk. 

\subsection{Speed Up}

In the speed up state, there is no need to follow an exact speed profile as the vehicle is trying to exit the crosswalk area slightly faster. In this case, the commanded acceleration is $a = a_\mathrm{cmf}$ until the pedestrian clears the crosswalk or once the vehicle passes the stopping location (i.e. $d< 0$). 

\section{Evaluation Methodology and Comparison with POMDP}

\subsection{Evaluation Methodology}

To validate the algorithm, both simulation and experimental studies were undertaken. The simulation results focused on testing Algorithm 1 against the specified design requirements over many pedestrian crossings. 

\begin{figure}
\centering
\includegraphics[width=2.0in]{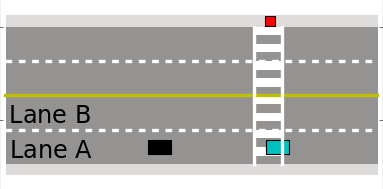}
\caption{Crosswalk simulation environment for algorithm validation. Ego vehicle is shown in black, leading vehicle in traffic stream shown in blue, and pedestrian shown in red.}
\label{fig:simFramework}
\end{figure}

The algorithm is validated in simulation via a built-from-scratch crosswalk simulation environment developed in Python (Fig.~\ref{fig:simFramework}). The simulation environment is built on the ROS robotics interface to enable easy porting of the control architecture to the experimental vehicle in Section \ref{sec:expres}, and for ease of comparison to the method provided by Thornton \cite{Thornton2018}.  

To simulate random pedestrian crossing events, a pedestrian is spawned on the sidewalk near the entrance to the crosswalk. The pedestrian accepts a randomly sized time gap in the traffic stream to cross the road at constant speed $\dot{x}_p$. The size of the time gap is drawn from a normal distribution with mean $\mu_{gap}$ and standard deviation $\sigma_{gap}$. Statistics for $\mu_{gap}$ and $\sigma_{gap}$ are obtained from results of the video graphic analysis by Feliciani et al. \cite{Feliciani2017}. The ego vehicle's response to the pedestrian is simulated over Algorithm 1 for the parameters shown in Table \ref{tb:params}. 

\begin{table}[h]
\begin{center}
\caption{Simulation and Controller Parameters}\label{tb:params}
\begin{tabular}{lccc}
Parameter & Symbol & Value & Units \\\hline\hline
Number of lanes & $n$ & 4 & - \\
Gap variance & $\sigma^2_{gap}$ & 2.5 & sec\\
Pedestrian velocity & $\dot{x}_p$ & 1.2 & $\mathrm{m/s}$ \\
Mean accepted gap & $\mu_{gap}$ & 4.0 & sec\\\hline
Safety offset    & $\Delta$ & 5.0 & m \\
Speed feedback gain & $k_s$ & 2.0 & 1 / sec\\
Brake time delay  & $t_\mathrm{delay}$ & 0 & sec \\ 
Speed limit & $v_\mathrm{speedlimit}$ & 4.5 & m/s \\
Comfort acceleration & $a_\mathrm{cmf}$ & 2 & $\mathrm{m/s^2}$ \\
Maximum time advantage & $\tau_\mathrm{max}$ & 4 & sec \\
Maximum acceleration & $a_\mathrm{max}$ & 9 & $\mathrm{m/s^2}$ \\\hline
\end{tabular}
\end{center}
\end{table}

\subsection{Alternate Approach - POMDP}
The method proposed by Thornton \cite{Thornton2018} offers a comparison to the proposed algorithm and models the unsignalized crosswalk problem as a Partially 
Observable Markov Decision Process (POMDP) \cite{Kochenderfer2015}. The relevant discretized state variable $x$ is given by

\begin{equation}
x_t = [v_t \hspace{2mm} c_t \hspace{2mm} d_t \hspace{2mm} p_t \hspace{2mm} a_{t-1}]^T
\end{equation}

Where $c_t$ is a Boolean variable that is true when the pedestrian is in the crosswalk, $a_{t-1}$ is the previous acceleration command, and $p_t$ is the pedestrian posture, which can be \textit{distracted}, \textit{walking}, or \textit{stopped}. Using a point mass model for the vehicle state transition dynamics and empirical data for the $c_t$ dynamics, the authors compute a closed loop policy $a_t = \pi(x_t)$ using the QMDP solver \cite{Kochenderfer2015}. The reward function that is optimized is given by: 

\begin{equation}
r(x_t, a_t) = g_\mathrm{legality} + g_\mathrm{safety} + g_\mathrm{efficient} + g_\mathrm{smooth}
\end{equation}

Where $g_\mathrm{legality}$ and $g_\mathrm{safety}$ are terms encouraging the vehicle to stop when the pedestrian is in the crosswalk, $g_\mathrm{efficient}$ is a term encouraging high velocity when there is no pedestrian, and $g_\mathrm{smooth}$ encourages the vehicle to avoid abrupt changes in acceleration. For our implementation of \cite{Thornton2018}, we assumed $p_t$ was set to \textit{stopped}. 

\section{Simulation Results}

\begin{figure}
\centering
\includegraphics[width=3.5in]{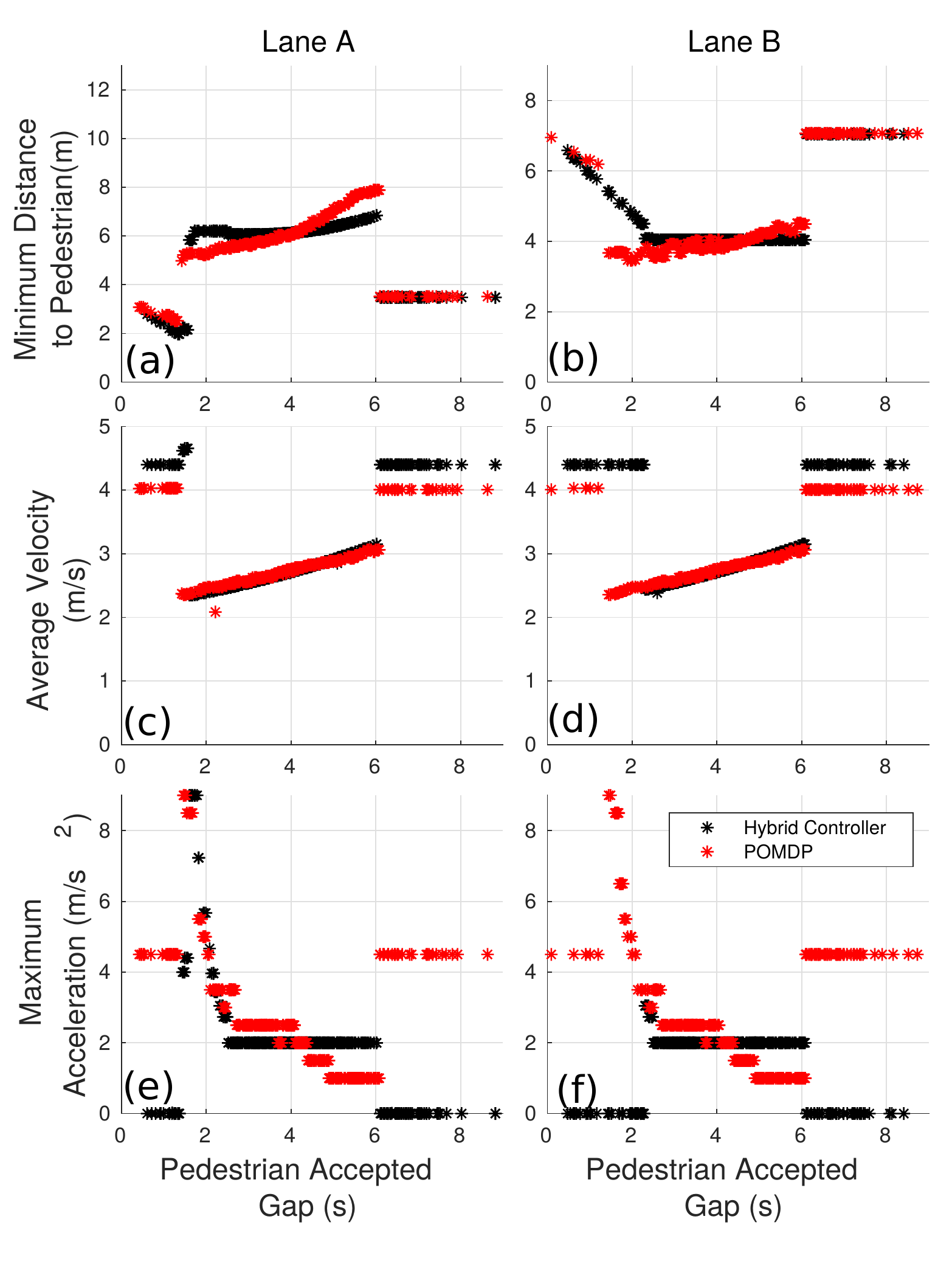}
\caption{Simulation results for case where pedestrian enters the crosswalk on the same side of the road as the vehicle. Plots (a) and (b) show the closest distance between the vehicle and pedestrian for each crossing trial, (c) and (d) plots the average vehicle velocity for each simulated crossing, and (e) and (f) plots the peak acceleration / deceleration magnitude for each trial.}
\label{fig:velPlot1}
\end{figure}

\begin{figure}
\centering
\includegraphics[width=3.5in]{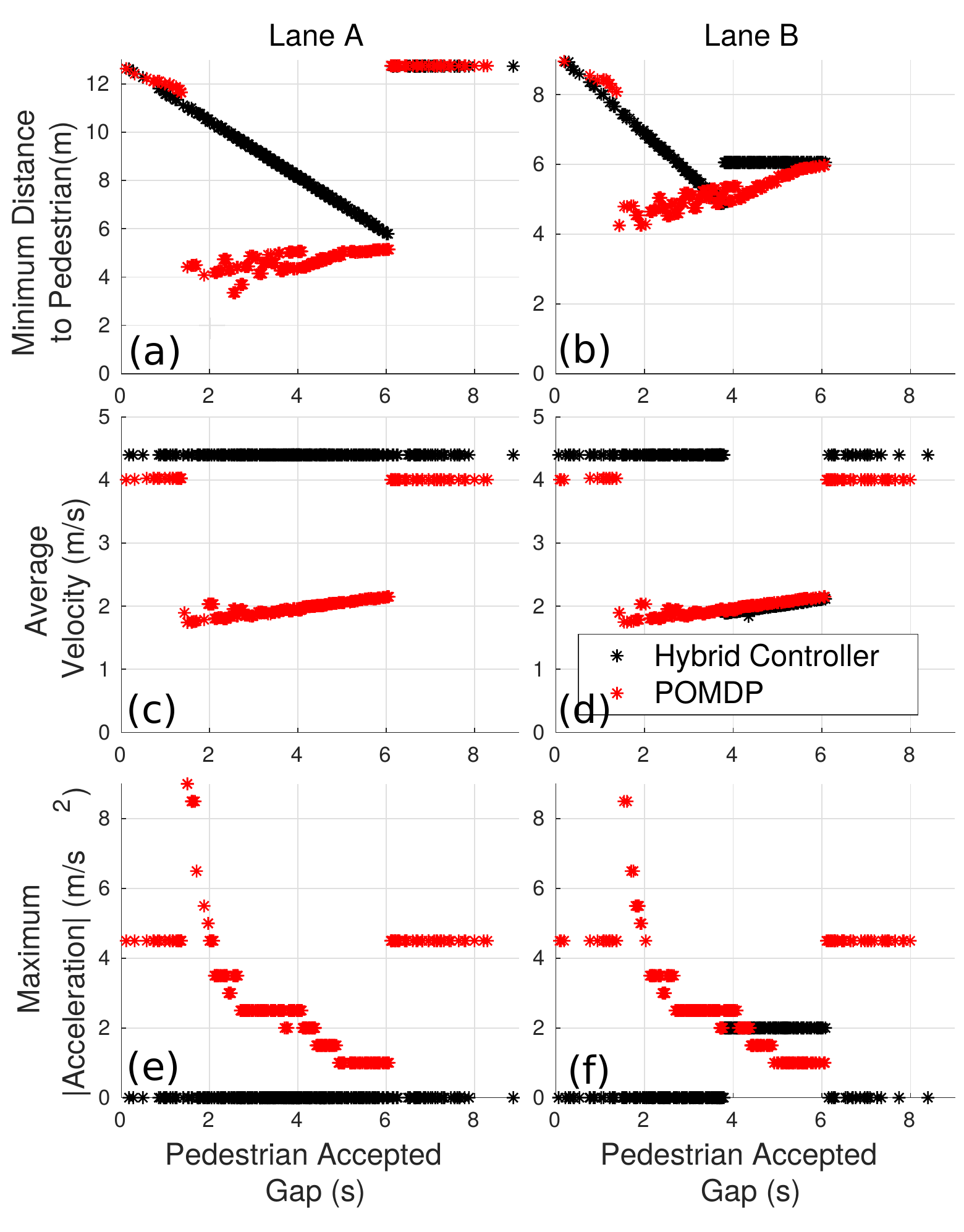}
\caption{Simulation results for case where pedestrian enters the crosswalk on the opposite side of the road as the vehicle. Plots (a) and (b) show the closest distance between the vehicle and pedestrian for each crossing trial, (c) and (d) plots the average vehicle velocity for each simulated crossing, and (e) and (f) plots the peak acceleration / deceleration magnitude for each trial.}
\label{fig:velPlot2}
\end{figure}

Figure \ref{fig:velPlot1} shows simulation results for the case where the pedestrian enters the crosswalk on the same side of the road as the vehicle. The vehicle is simulated in both Lane A and Lane B in Fig.~\ref{fig:simFramework}. Every marker in the figure represents one of 1500 simulated vehicle crossings, 750 for each method. Figure \ref{fig:velPlot2} shows the same simulation results, but for the case where the pedestrian enters the crosswalk on the opposite side of the road as the vehicle.

\subsection{Safety and Legality}

To meet the collision avoidance design requirement, we generated enough simulations to ensure errant pedestrian crossing behavior at ``risky" accepted gaps ($<$~3 seconds). The minimum distance between the vehicle and the pedestrian for each trial is shown in (a) and (b) of Figures ~\ref{fig:velPlot1} and \ref{fig:velPlot2}. The vehicle is able to maintain a safe distance of at least four meters from the pedestrian for the case where the vehicle is Lane B. This is because the algorithm has enough time to react to a pedestrian crossing no matter what size gap is accepted. 

The riskiest case occurs when the vehicle is in Lane A and the pedestrian crosses from the right side, accepting a gap between 1.25 and 1.75 seconds. This is unlikely given gap acceptance statistics reported in the literature \cite{Rasouli}\cite{Feliciani2017}, but would represent a safety risk given the vehicle has limited time to react. However, since the algorithm recognizes that braking will not clear the crosswalk, the vehicle is able to speed up and maintain at least 2 meters of lateral distance from the pedestrian at all times. Future work will use formal verification methods in order to show safety for all pedestrian accepted gaps. 


\subsubsection{Comparison with POMDP}

In terms of safety and legality, both the POMDP and hybrid controller maintain safe distances from the pedestrian across all trials. For the hybrid controller, this is because the need to stop at $d=0$ is explicitly encoded in the design. For the POMDP, this is due to the reward function terms $g_\mathrm{legality}$ and $g_\mathrm{safety}$ 

\subsection{Efficiency}

 Figures ~\ref{fig:velPlot1} and ~\ref{fig:velPlot2} show the average velocity of the vehicle for each simulated crossing in (c) and (d). For high accepted gaps, the pedestrian waits for the ego vehicle to exit the crosswalk before beginning to walk, and the algorithm does not need to slow down from the speed of traffic. For low accepted gaps, the vehicle cannot stop, and must speed up or continue driving through the intersection, also keeping the vehicle speed high. 

The most disruption to traffic flow occurs when the vehicle must slow down significantly or stop, which occurs for accepted gaps between 2-6 seconds. A special case occurs in Fig.~\ref{fig:velPlot2} where the vehicle is in Lane A and the pedestrian crosses from the opposite side of the road. Because the vehicle is at a far distance from the pedestrian, the time advantage remains sufficiently high for the vehicle to comfortably continue through the crosswalk without needing to slow down. 

\subsubsection{Comparison with POMDP}
The policy from the POMDP displays interesting behavior at low and high pedestrian accepted gaps. At high accepted gaps, even though the pedestrian is waiting for the car to pass before crossing, the POMDP still slows the car down. The POMDP, which uses a probabilistic model of pedestrian crossing, is accounting for the chance the pedestrian might cross at a very low accepted gap. The POMDP policy does not account for the possibility of the vehicle speeding up at the last second. The hybrid control policy has this case built in, and does not need to slow down if the pedestrian has not entered the crosswalk, resulting in an overall efficiency gain. 

Also note the POMDP models the pedestrian as being in the crosswalk or out of the crosswalk - this is necessary to limit the size of the discretized state space for the QMDP solver. This results in the POMDP having the same control policy across all four simulated cases. Because the proposed hybrid controller is continuous, there is no computational burden on accounting for the position and velocity of the pedestrian in the crosswalk. This allows for the controller to consider factors such as time advantage, and continue through the crosswalk in many cases. 





\subsection{Smoothness}
Plots (e) and (f) of Figures~\ref{fig:velPlot1} and \ref{fig:velPlot2} show the peak acceleration / deceleration magnitude for each simulated crossing. The hybrid controller maintains a peak acceleration of 2 $\mathrm{m/s^2}$ for the vast majority of simulated trials, particularly when the pedestrian crosses from the opposing side of the road or when the vehicle is in Lane B.

However, there are a handful of simulated trials in Fig.~\ref{fig:velPlot1} when the pedestrian accepts a gap between 1.75-2.5 seconds and the vehicle is in Lane A. In this case, the vehicle must enter the \textit{Hard Braking} state and decelerate at a magnitude higher than $a_\mathrm{cmf}$. This is required in order to meet the collision avoidance constraint, which naturally takes priority.

\subsubsection{Comparison with POMDP}
 One relative advantage of the POMDP is that it is able to decelerate at rates below $a_\mathrm{cmf}$. At pedestrian accepted gaps of 5-6 seconds, the vehicle decelerates early and at a very low rate in order to avoid excessive jerk. However, by using the same policy for all four simulated scenarios and by not considering the ability to drive through the intersection, the POMDP also tends to decelerate when driving through the intersection would have sufficed. 

\begin{figure}[h]
\centering
\includegraphics[width=3.0in]{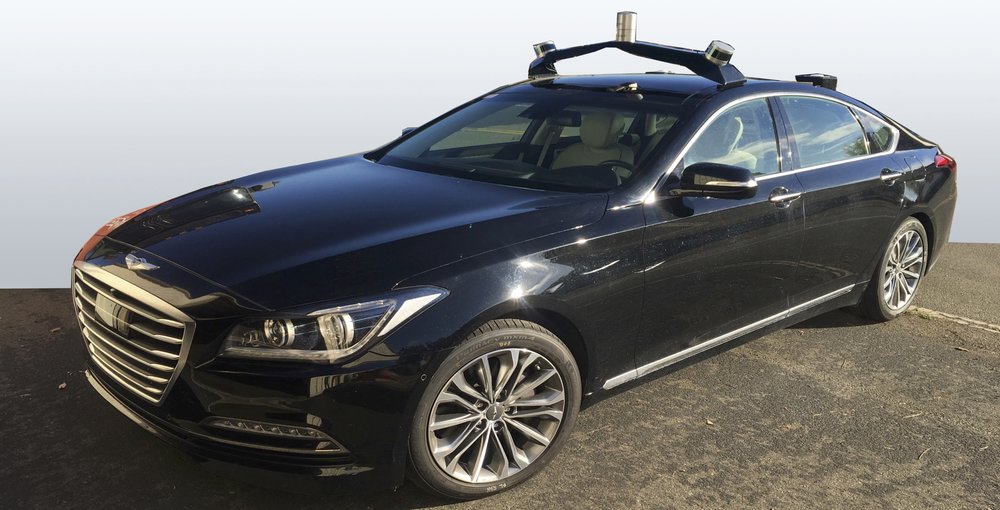}
\caption{Hyundai Genesis G80 experimental testbed.}
\label{fig:g80}
\end{figure}

\begin{figure}[h]
\centering
\includegraphics[width=2.0in]{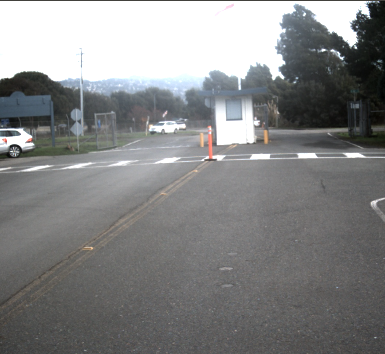}
\caption{Two lane crosswalk used for experimental testing.}
\label{fig:crosswalkpic}
\end{figure}

\begin{table}[h]
\begin{center}
\caption{Experimental Parameters}\label{tb:expparams2}
\begin{tabular}{llll}
Trial Number & Gap Accepted (sec) & Pedestrian Entry & Mode\\\hline\hline
1 & 4.0 & right & Yield\\ 
2 & 1.0 & right & Speed Up\\ 
3 & 7.0 & right & Yield\\ 
4 & 2.5 & right & Hard Brake\\ 
5 & 3.0 & left & Yield \\ 
6 & 1.0 & left & Speed Up\\\hline
\end{tabular}
\end{center}
\end{table}

\section{Experimental Results}
\label{sec:expres}

The proposed hybrid controller is tested experimentally using a Hyundai Genesis G80 experimental testbed, shown in Fig. \ref{fig:g80}. A two lane unsignalized intersection near the Berkeley campus was chosen for the experimental validation. In lieu of an actual pedestrian, a pedestrian was simulated virtually using the ROS simulation architecture, with the position and velocity of the pedestrian passed to the controller. Differential GPS was used to record the distance to the crosswalk, and a steering controller from \cite{Kapania2015} was used to autonomously steer the vehicle down the road. Relevant parameters of the experiments are shown below in Table \ref{tb:expparams}. The experiment consisted of six trials intended to show operation of the four discrete modes of the controller. A description of the trial parameters are shown in Table \ref{tb:expparams2}. 

\begin{table}[h]
\begin{center}
\caption{Experimental Parameters}\label{tb:expparams}
\begin{tabular}{lccc}
Parameter & Symbol & Value & Units \\\hline\hline
Number of lanes & $n$ & 2 & - \\
Safety offset  & $\Delta$ & 5 & m \\
Speed feedback gain & $k_s$ & 1.0 & 1 / sec\\
Brake time delay  & $t_\mathrm{delay}$ & 0.5 & sec \\ 
Speed limit & $v_\mathrm{speedlimit}$ & 7 & m/s \\
Comfort acceleration & $a_\mathrm{cmf}$ & 2 & $\mathrm{m/s^2}$ \\
Maximum time advantage & $\tau_\mathrm{max}$ & 4 & sec \\
Maximum acceleration & $a_\mathrm{max}$ & 9 & $\mathrm{m/s^2}$ \\\hline
\end{tabular}
\end{center}
\end{table}

\begin{figure}[h]
\centering
\includegraphics[width=3.5in]{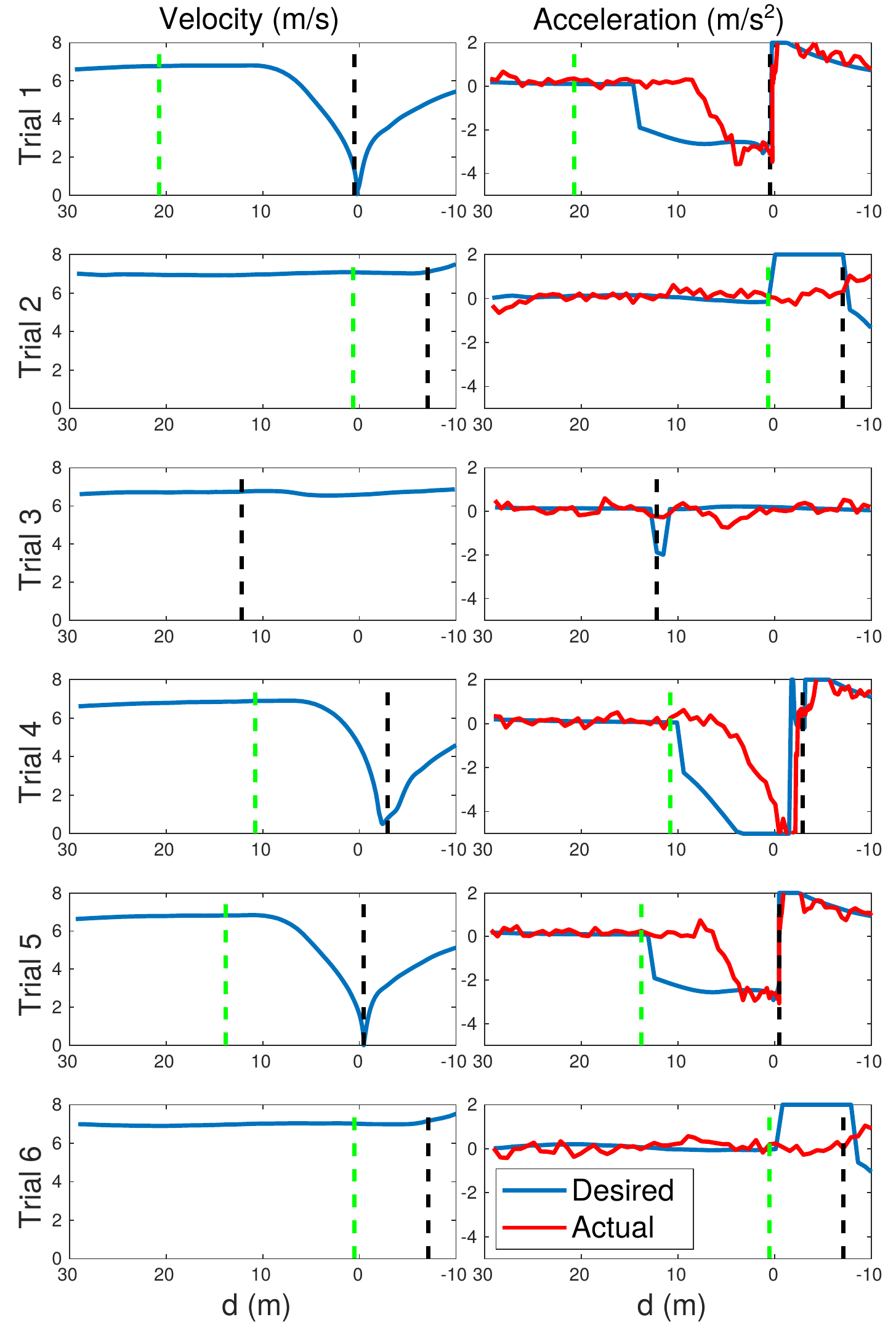}
\caption{Experimental velocity and acceleration plotted vs distance to crosswalk. Green lines mark the point where the pedestrian first enters crosswalk. Black lines mark the point where the pedestrian leaves the crosswalk. Note that for Trial 3, the green line is off the page at 40 meters.}
\label{fig:expPlot}
\end{figure}

The experimental velocity and acceleration is plotted vs $d$ in Figure \ref{fig:expPlot}. In Trial 1, the pedestrian accepts a gap of four seconds, giving the vehicle enough time to yield at a deceleration of 2 $\mathrm{m/s^2}$. Note that the low level longitudinal controller exhibits a large time delay between commanded and actual acceleration. Because this is accounted for in Algorithm 1, the vehicle is able to brake early and the vehicle is stopped right at $d=0$ before the pedestrian exits the crosswalk and the vehicle speeds back up. 

In Trial 2, the pedestrian enters from the right with just 1 second before collision. This experiment was undertaken to verify the vehicle would not brake, but accelerate through the intersection. Trial 3 represents a situation where the pedestrian crosses with a large (7 second) gap. In this case, the pedestrian has enough time to almost cross before the vehicle needs to yield, resulting in the vehicle only needing to slow down slightly. 

In Trial 4, the pedestrian exhibits risky behavior, crossing with a 2.5 second gap. The vehicle has enough time to brake, but must decelerate rapidly at 5 $\mathrm{m/s^2}$ in the \textit{hard braking} mode. Given the brake delay in the experimental vehicle, significant overshoot occurs, and the vehicle stops nearly 2.5 meters ahead of the desired stopping location. However, given that the desired stop location is 5 meters ahead of the crosswalk, this is sufficient to avoid a collision.

Trials 5 and 6 represent pedestrian crossings from the opposite side of the road under accepted gaps of 3.0 and 1.0 seconds, respectively. In the former case, the vehicle is able to yield in a fashion similarly to Trial 1, and in the latter case, the vehicle is able to speed through the intersection, as in Trial 2. 

\section{CONCLUSIONS}
This paper proposes a novel hybrid control architecture for determing closed loop vehicle control at an unsignalized intersection. Through simulation, we show that the system ensures safety across a continuous spectrum of pedestrian gap acceptance behaviors, while balancing the competing demands of smoothness for the vehicle occupants and limited traffic slowdown for other vehicles in the traffic stream. Remarkably, with just four distinct modes, the controller is able to handle multiple lanes and pedestrians crossing from either side of the road. Further work will focus heavily on integrating perception to accurately determine when to trigger switches in the controller modes, as determining pedestrian intent to cross is a major input required for the controller. 

\addtolength{\textheight}{-12cm}  





\bibliographystyle{IEEEtran}
\bibliography{Bibliography}

\end{document}